# An Autonomous Path Planning Method for Unmanned Aerial Vehicle based on A Tangent Intersection and Target Guidance Strategy

Huan Liu, Xiamiao Li, Mingfeng Fan, Guohua Wu,

Witold Pedrycz, *Fellow*, *IEEE*, Ponnuthurai Nagaratnam Suganthan, *Fellow*, *IEEE*

*Abstract*—Unmanned aerial vehicle (UAV) path planning enables UAVs to avoid obstacles and reach the target efficiently. To generate high-quality paths without obstacle collision for UAVs, this paper proposes a novel autonomous path planning algorithm based on a tangent intersection and target guidance strategy (APPATT). Guided by a target, the elliptic tangent graph method is used to generate two sub-paths, one of which is selected based on heuristic rules when confronting an obstacle. The UAV flies along the selected sub-path and repeatedly adjusts its flight path to avoid obstacles through this way until the collision-free path extends to the target. Considering the UAV kinematic constraints, the cubic B-spline curve is employed to smooth the waypoints for obtaining a feasible path. Compared with A∗, PRM, RRT and VFH, the experimental results show that APPATT can generate the shortest collision-free path within 0.05 seconds for each instance under static environments. Moreover, compared with VFH and RRTRW, APPATT can generate satisfactory collision-free paths under uncertain environments in a nearly real-time manner. It is worth noting that APPATT has the capability of escaping from simple traps within a reasonable time.

## I. INTRODUCTION

IN recent years, unmanned aerial vehicles (UAVs) have received significant attention, and are extensively employed in various fields, such as traffic inspection [1], [2], disaster rescue [3], cargo delivery [4], [5], and target reconnaissance [6]. It is worth noting that path planning plays a key role to realize the autonomous control of the UAV system, which facilitates the effective application of UAVs. Currently, it is still a significant challenge to realize efficient path planning for UAVs in complex environments with dense obstacles and uncertainties while considering several demands, such as obstacle avoidance, trajectory feasibility, real-time planning capability, and satisfactory path length.

Quite a few methods have been proposed to generate collision-free paths for UAVs in obstacle environments. Path planning methods can be roughly divided into two types: graph theory based and non-graph theory based methods. The graph theory based approaches aim to find a reasonable path in a certain graph modeling the environment, where the A∗ algorithm is known to be effective at finding a desirable path to a target while avoiding obstacles [7]. Classical non-graph theory based approaches include rapidly exploring random tree [8], vector field histogram [9], genetic algorithm [10], which plan a path based on random sampling, potential filed theories or bionics theories.

Nevertheless, the most existing approaches based on graph theory are time-consuming on account of the entire roadmap generation and cannot satisfy the real-time path planning requirement. Besides, traditional approaches are hard to make a good tradeoff between time efficiency and solution quality, especially in the complex environment of dense obstacles and uncertainties [10], [11].

To plan collision-free paths efficiently in the static or dynamic environments, this paper proposes a novel autonomous path planning algorithm based on a tangent intersection and target guidance strategy (APPATT). Guided by a target, the path departs from a start-point and circumnavigates obstacles based on the elliptic tangent graph. To search satisfactory collision-free paths, APPATT comprehensively takes into account the conditions (i.e., the estimated path length and obstacle avoidance) of local paths from a waypoint to the

This work was supported by the National Natural Science Foundation of China under Grant 61603404 and Natural Science Fund for Distinguished Young Scholars of Hunan Province under Grant 2019JJ20026.

Huan Liu is with the School of Traffic and Transportation Engineering, Central South University, Changsha 410075, China. E-mail: LiuHuan1095@csu.edu.cn.

Xiamiao Li is with the School of Traffic and Transportation Engineering, Central South University, Changsha 410075, China. E-mail: xmli@csu.edu.cn.

Mingfeng Fan is with the School of Traffic and Transportation Engineering, Central South University, Changsha 410075, China. E-mail: mingfan@csu.edu.cn.

Guohua Wu *(the corresponding author)* is with the School of Traffic and Transportation Engineering, Central South University, Changsha 410075, China. E-mail: guohuawu@csu.edu.cn.

Witold Pedrycz is with the Department of Electrical and Computer Engineering, University of Alberta, Edmonton, AB T6G 2V4, Canada, with the Department of Electrical and Computer Engineering, Faculty of Engineering, King Abdulaziz University, Jeddah 21589, Saudi Arabia, and also with the Systems Research Institute, Polish Academy of Sciences, Warsaw 01447, Poland. E-mail: wpedrycz@ualberta.ca.

Ponnuthurai Nagaratnam Suganthan is with the School of Electrical Electronic Engineering, Nanyang Technological University, Singapore, 639798. E-mail: EPNSugan@ntu.edu.sg.

start-point and to the target.

The main contributions of this paper are summarized as follows:

1) We propose a novel graph-based path planning algorithm APPATT, which does not need to generate the entire roadmap. Two sub-paths are expanded based on the elliptic tangent graph when confronting an obstacle. Meanwhile, one of the two sub-paths is selected based on four heuristic rules. The UAV flies along with the selected sub-path and repeatedly adjusts its flight path to avoid obstacles through this way until it arrives at the target successfully. It can be found that in APPATT, only one collision-free path would be recorded on the map instead of many candidate paths on the map traditionally.

2) The APPATT can also realize real-time path planning in dynamic environments. The main thought of APPATT in dynamic environments is similar to that in static environments. The difference is that APPATT in dynamic environments limits the flight distance between two adjacent waypoints which can increase the number of waypoints to perceive the environment frequently based on sensors. In partially or completely unknown environments, the proposed APPATT can generate collision-free paths based on real-time perceived information with lower computational complexity.

3) We conduct extensive experiments in static and dynamic environments to validate the effectiveness of the proposed APPATT. The experimental results demonstrated that APPATT could plan a high-quality collision-free path efficiently in the static environments with dense obstacles. In particular, APPATT could escape from simple mazes in a reasonable time. Furthermore, APPATT could respond to dynamic changes rapidly and realize real-time path planning.

The rest of this paper is organized as follows. Section II reviews the related works. Section III states the path planning problem. Section IV presents the proposed algorithm in detail. Sections V reports the computational and comparative results. Finally, conclusions are drawn in Section VI.

## II. RELATED WORKS

Path planning is one of the fundamental tasks for the UAV operation. It can be simply described as seeking a feasible, collision-free and optimal path between two positions [12]. So far, four types of path planning methods have been proposed, namely graph-based methods, sampling-based methods, potential field methods and intelligent optimization methods [13]. Classical graph-based methods include the Voronoi diagram, the visibility graph, the A* algorithm, the tangent graph and so on. The Voronoi diagram [14] approximately partitions the workspace into obstacle-centered free cells, which improves the efficiency of path planning. The paths on the Voronoi diagram are far from obstacles, whereas the path length is not guaranteed to be optimal [15]. The visibility graph [16] can easily identify a satisfactory path, but it must reconstruct the roadmap once the start-point or the end-point change. The A* algorithm [17] is a popular path planning algorithm and has the capability to escape from mazes. The three methods above are reliant on the global environment so that once the environment changes a little, the entire path must be re-planed. Hence, they are usually employed in static environments and cannot be directly exploited to perform the UAV path planning in dynamic environments [18], [19].

Rohnert [20] proposed an algorithm (i.e., classical tangent graph) to find a desirable path among convex polygons by using common tangents of the polygons. The classical tangent graph needs expensive computational resources to store common tangents of the polygons, and the path is tortuous and close-to-obstacle. To overcome these shortcomings, Chen *et al*. [21] enclosed obstacles in circles. Nevertheless, the circular enveloping efficiency is not high enough, and it is easy to excessively transform the feasible area into the infeasible area, resulting in a waste of free space. Petillot *et al*. [22] and Liu *et al*. [23] enclosed obstacles in ellipses which could describe narrow obstacles and corridors well. It is worth noting that the improved tangent graph still needs to construct the entire roadmap prior to the path planning operation, which increases the burden of storage. In conclusion, existing tangent graph based methods have the following drawbacks: (1) they relatively lower time efficiency in the path planning due to the generation of the entire roadmap. (2) They are not suitable to real-time path planning scenarios.

Sampling-based methods, such as the probabilistic roadmap method (PRM) [24] and the rapidly exploring random tree (RRT) [8], show superiority in path exploration, which renders them among successful methods for UAV path planning. PRM has multiple two-point boundary value problems during the roadmap construction [11], which leads to high computational costs and poor performance in dynamic environments. On the contrary, RRT does not need to sample the space and construct the roadmap before path planning [25]. RRT has a powerful spatial search ability and works efficiently in complex and dynamic environments. Owing to its randomness in nature, classical RRT involves no mechanism for an improvement of the quality of the path so that its performance is not satisfactory in static environments [26].

In the field of online obstacle avoidance, potential field algorithms, such as artificial potential field (APF) [27], Bug algorithm [28] and vector field histogram (VFH), have attracted much attention. To deal with the oscillation in APF [29], Borenstein and Koren [9] developed the vector field histogram (VFH), a reactive method that looks for gaps in constructed polar histograms of the UAV's location. The VFH algorithm is robust and can plan a collision-free path in near real time. Based on the VFH, VFH$^+$ [30] and VFH* [31] were further developed in which some factors, such as the robot's width and available trajectories are considered. Intelligent optimization algorithms, including genetic algorithm (GA) [10], ant colony algorithm (ACO) [19], [32], particle swarm optimization (PSO) [33], etc., play an important role in path planning in complex environments. They mostly have global optimization capability, but they are generally time-consuming especially in densely obstructive environments and have poor performance in real-time path planning.

From the survey mentioned above, it can be found that: (1) major graph-based methods are time-consuming in complex environments and not suitable to real-time path planning as they need to construct the entire roadmap. (2) It is still a significant challenge for UAV path planning to achieve a balance between time efficiency and solution quality.

In brief, this paper proposes a novel graph-based algorithm named APPATT for short. The proposed APPATT can be applied to both offline and online scenarios with polygonal shapes obstacles represented by ellipses.

## III. PROBLEM STATEMENT

We take a scenario in Fig. 1 as an example. Suppose that a UAV delivers goods from the start-point $S$ to the end-point $E$. Since high-rise buildings in the city prevent the UAV from flying directly from $S$ to $E$, it is necessary to plan a desirable collision-free path between two positions. To reduce computation time and narrow the path search space, the three-dimensional scenario can be simplified into the two-dimensional scenario, as shown in Fig. 2. It is reasonable that we focus on two-dimensional scenarios from the planning level while three-dimensional scenarios would be considered from the operational level. The shapes of polygons are not uniform, which may lead to some problems such as expensive computation, unsmooth and close-to-obstacle paths. In order to overcome the abovementioned shortcomings, obstacles are uniformly modeled as ellipses, which can describe various obstacles and facilitate successive path planning operations.

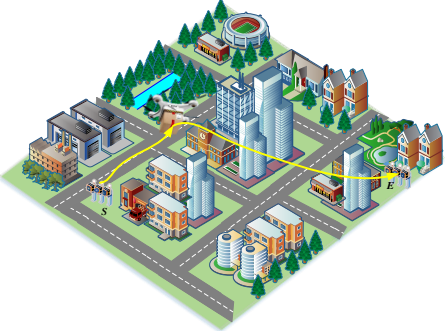

Fig. 1 UAV delivery scenario

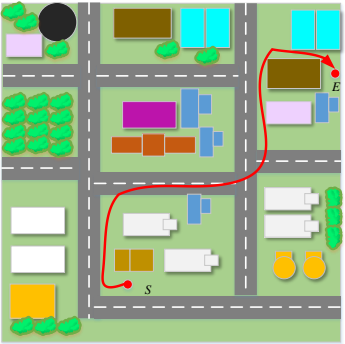

Fig. 2 Two-dimensional scenario

Suppose that the start-point is $S$ and the end-point is $E$. There are $N$ obstacles in the environment. A set of obstacles is represented as $B = \{B_1, B_2, \cdots, B_k, \cdots, B_N\}$, and the center coordinate of the obstacle $B_k$ is denoted as $(x_k, y_k)$. The symbols $a$ and $b$ are the two semi-axis of the ellipse, and $\theta$ is the inclination of semi-major axis. $r_{safe}$ is the minimum safe distance required between a UAV and an obstacle, which ensures that the path is far enough from the obstacle. The obstacle can be denoted as

$$\frac{[(x-x_k)\cos\theta+(y-y_k)\sin\theta]^2}{(a+r_{safe})^2} + \frac{[(y-y_k)\cos\theta-(x-x_k)\sin\theta]^2}{(b+r_{safe})^2} = 1. \quad (1)$$

Thus, each waypoint $P_i(x_i, y_i)$ on the collision-free path should meet the following formula.

$$\frac{[(x_i-x_k)\cos\theta+(y_i-y_k)\sin\theta]^2}{(a+r_{safe})^2} + \frac{[(y_i-y_k)\cos\theta-(x_i-x_k)\sin\theta]^2}{(b+r_{safe})^2} \geq 1. \quad (2)$$

where $P_i$ is a waypoint generated by the proposed APPATT, and $i = 1, 2, \cdots, n$; $k = 1, 2, \cdots, N$. The edge comprised of any two adjacent waypoints is a tangent of the ellipse encapsulating the obstacle, which ensures that the edge is kept a safe distance from obstacles.

The UAV path can be described as follows: the UAV may start from the start-point $S$ and pass several waypoints generated based on the elliptic tangent graph until it reaches the end-point without obstacle collision. The set of waypoints is given as (3):

$$Route = \{S, P_1, P_2, \cdots, P_n, E\}. \quad (3)$$

Path planning must meet platform constraints of the UAV, mainly including:

*1) Maximum range constraint.* The range of the UAV is limited by fuel quantity during the entire flight. $l_i$ is the flight distance of the $i$-th sub-path. $L_{max}$ is the maximum range. Suppose there are $m$ waypoints. The flying range of UAV must meet the following formula.

$$L \leq L_{max}, L = \sum_{i=1}^{m-1} l_i. \quad (4)$$

*2) Minimum route leg length.* This constraint restricts the path to be straight for a predetermined minimum distance before initiating a turn [34].

*3) Minimum turning radius.* The UAV's heading can be varied by adjusting the rudder and aileron angles. However, due to the inertial effect, change of heading requires time and turning radius, so the minimum turning radius needs to be considered in path planning.

## IV. DESIGN OF APPATT

Traditional graph-based path planning methods mainly include two steps [35]: the first step is to establish a roadmap and the second step is to search for a high-quality path on this map. To search for a high-quality path in the second step, they usually generate many candidate collision-free paths from a start-point to a target on the roadmap in the first step. A large-scale roadmap not only increases the storage but also obviously decreases search efficiency.

Instead of constructing the entire roadmap, when confronting an obstacle, APPATT generates two sub-paths based on the elliptic tangent graph. One of the two sub-paths is selected according to sophisticatedly designed rules considering the obstacle avoidance condition and the sub-path length. The UAV flies along the selected sub-path and repeatedly adjusts its flight path to avoid obstacles through this way until the collision-free path extends to the target. Only a collision-free path will be retained on the map. According to whether the environments of application scenarios are known in advance, we design two APPATT versions: static elliptic tangent graph method based on target guidance (SETG-TG) and dynamic elliptic tangent graph method based on target guidance (DETG-TG). In the SETG-TG algorithm, all the information about the environment is supposed to be known in advance. By contrast, sometimes limited knowledge about the environment is available, thus the DETG-TG algorithm relies on the information gathered by the sensors in real time.

The objective of APPATT is to produce a high-quality path for a UAV efficiently [36]. The notations used many times to describe APPATT are listed in Table I. In order to describe the algorithm clearly in this section, we define the tangents of an obstacle (i.e. an ellipse modeling the obstacle) from the point $O$ or the point $D$ as origin-tangents or destination-tangents, respectively. Except for a start-point and an end-point, all points in the set $Pa$ and $Ca$ stem from intersection points of origin-tangents and destination-tangents. A collision-free path can be generated by connecting the waypoints in the set $Pa$ sequentially, while the paths from the waypoints in the set $Ca$ to the point $O$ are infeasible. For example, two intersection points can be generated by drawing two tangents of an obstacle from $O$ and $D$ respectively, one of which is selected as a waypoint based on heuristic rules. If the straight path from the waypoint to $O$ (i.e. an origin-tangent) is collision-free, the waypoint will be added to the set $Pa$; otherwise, it will be added to the set $Ca$.

TABLE I
DEFINITIONS OF MAIN NOTATIONS

| Notations | Description |
|---|---|
| $S$ | start-point |
| $E$ | end-point |
| $T$ | a waypoint generated by APPATT |
| $O, D$ | origin and destination points, i.e., two vertices of each sub-path (these two points are updated during the path planning process) |
| $Pa$ | a set of determined waypoints used to generate collision-free edges in order |
| $Ca$ | a set of candidate waypoints that may contribute to the final path |
| $Ba$ | a set that records the obstacles avoided in sequence |

### A. SETG-TG Algorithm

#### 1) The main procedure of the SETG-TG Algorithm

To describe the core idea of the SETG-TG algorithm well, we especially explain the following terms. (1) Obstacles: the ellipses modeling the obstacles in the text below. (2) Start-point: the location where the UAV takes off. (3) End-point: the location where the UAV finally reaches. (4) Waypoint: it derives from intersection points of origin-tangents and destination-tangents. (5) Origin: the UAV's location. An origin stems from a start-point or a waypoint. The waypoints that precedes an origin can be connected into a collision-free path. The origin can change dynamically during the UAV path searching process. (6) Destination: the point that the UAV should temporarily fly to. It may be an end-point or sometimes a waypoint. (7) First-collided obstacle: the obstacle that collides with the straight path from an origin and a destination and is closer to the origin. (8) Sub-path: a temporary path composed of an origin-tangent and a destination-tangent. It starts from an origin and ends at a destination.

Generally, collision-free paths can be generated via two steps: waypoint generation and the update of the origin and the destination. To be specific, firstly, a UAV departs from a start-point and takes the start-point and end-point as the origin and the destination, respectively. The straight path from the origin to the destination can be infeasible in the presence of obstacles. Two temporary paths (can be feasible or infeasible sub-paths) are generated by drawing two tangents of the first-collided obstacles from the origin and the destination respectively, one of which is selected in accordance with several effective heuristic rules, which will be detailed later. For instance, as shown in Fig. 3(a), the point $S$ is a start-point and the point $E$ is an end-point. We regard the point $S$ and $E$ as an origin and a destination respectively. The straight path $SE$ collides with the obstacle $B_1$ (i.e. the first-collided obstacle). We can generate two tangents of the obstacle $B_1$ from the point $S$, namely $SF_1$ and $SF_1'$. Similarly, two destination-tangents, namely $EF_1$ and $EF_1'$, are generated. Hence, two temporary paths (i.e. $S \to F_1 \to E$ and $S \to F_1' \to E$) are produced. The waypoint $F_1$ is selected according to heuristic rules.

Secondly, we should judge whether the straight path from the origin $S$ to the waypoint $F_1$ is collision-free. If it is true, the path extends to the waypoint $F_1$ and the waypoint $F_1$ is updated to be the origin. Otherwise, the waypoint $F_1$ is updated to be the destination. The above-mentioned operation to avoid obstacles is performed in an iterative manner until the final collision-free path to the end-point is obtained. As shown in Fig. 3(a), the straight path from the origin $S$ to the waypoint $F_1$ collides with the obstacle $B_5$, thus the waypoints $F_1$ is updated to be the destination. We can generate the waypoints between an origin and a destination via the elliptic tangent graph method iteratively until the collision-free path can extend to the end-point, namely $S \to F_3 \to F_1 \to F_2 \to E$.

The pseudocode of the SETG-TG algorithm is given in Algorithm 1. The main procedure of the SETG-TG algorithm is as follows:

*Step 1 (Obstacle modeling):* Follow the formula (1) to enclose the obstacles in ellipses (Line 1), as shown in Fig. 3.

*Step 2 (Initialization):* Add the start-point $S$ and the end-point $E$ to the waypoint set $Pa$ and the set of candidate waypoints $Ca$, respectively. The set $Ba$ recording the obstacles being avoided is initialized to an empty set (Line 2).

*Step 3 (Judging whether the path is collision-free):* Take the last points in the set $Pa$ and $Ca$ as $O$ and $D$, respectively (Line 4). Then, connect $O$ to $D$ and judge whether the path $OD$ is collision-free (Line 5). If true, add the point $D$ to the set $Pa$, delete the point $D$ from the set $Ca$ (Lines 6-7), and go to Step 6; otherwise, mark and record the first-collided obstacle and go to Step 4.

*Step 4 (Waypoint Generation):* Generate two tangents of the first-collided obstacle from $O$ and $D$ respectively. Two temporary paths (i.e. sub-paths) can be generated (Line 9).

When selecting one of the two sub-paths, the SETG-TG algorithm should take into account some factors, such as the estimated distance and obstacle avoidance condition of two sub-paths. In particular, the conditions of destination-tangents are taken into account as well, which can accelerate search speed and ensure the heading accuracy. Path generation is realized gradually by selecting one sub-path from two temporary paths when facing an obstacle, which follows the four priority rules in order: (1) we generate a set $Ba$ recording the obstacles being avoided each time. Choose the sub-path whose origin-tangent does not collide with the last obstacle in the set $Ba$, which can prohibit the UAV from going back to the start-point. (2) Choose the sub-path whose origin-tangent collides with less obstacles. (3) Choose the sub-path whose destination-tangent collides with less obstacles. (4) Choose the sub-path whose length is shorter. As shown in Fig. 3(a), the obstacle $B_1$ is recorded to the set $Ba$ in the first obstacle avoidance process. The origin-tangent (i.e., $SF_1'$) crashes against the obstacles $B_2$ and $B_6$ while $SF_1$ only crashes against the obstacle $B_5$. We cannot select one sub-path based on rule (1) because the obstacles $B_2$, $B_5$ and $B_6$ are not recorded in the set $Ba$. Consequently, the sub-path $S \to F_1 \to E$ is selected according to rule (2) and $F_1$ is a waypoint $T$. In addition, a UAV may fall into the maze-like environment as shown in Fig. 3(b). The origin-tangent $SF_2'$ collides with the obstacle $B_1$ being avoided in the first obstacle avoidance process. As a result, the point $F_2$ is selected as a waypoint $T$ according to rule (1).

*Step 5 (Judging whether the waypoint T is available directly):* Connect $O$ to $T$ and judge whether the path $OT$ is collision-free (Line 10). If true, add the waypoint $T$ to the set $Pa$ (Lines 11-12); otherwise, add the waypoint $T$ to the set $Ca$ (Line 14). Then, record the first-collided obstacle to the set $Ba$ (Line 16).

*Step 6 (Stop Criteria):* Go to Step 3 and repeat the above steps until the path visits the end-point without obstacle collision (Lines 3-18).

---

**Algorithm 1** SETG-TG

**Input:** Start-point $S$, End-point $E$;
**Output:** $Pa$;
1: Generate a minimum external ellipses for each obstacle according to the formula (1);
2: Initialize: $Pa \leftarrow \{S\}, Ca \leftarrow \{E\}, Ba \leftarrow \phi$;
3: **While** $Ca$ is not empty
4:     $O \leftarrow Pa(end), D \leftarrow Ca(end)$;
5:     Connect $O$ to $D$, judge whether the path $OD$ collides with obstacles;
6:     **If** the path $OD$ is collision-free
7:         Add $D$ to $Pa$, and delete $D$ from $Ca$;
8:     **Else**
9:         Generate two tangents of the first-collided obstacle from point $O$ and $D$ respectively, and obtain two sub-paths. Then, choose a better sub-path and update a waypoint $T$ according to four rules;
10:     Connect $O$ and $T$, judge whether the path $OT$ collides with obstacles;
11:     **If** the path $OT$ is collision-free
12:         Add the waypoint $T$ to $Pa$;
13:     **Else**
14:         Add the waypoint $T$ to $Ca$;
15:     **End if**
16:     Record the first-collided obstacle to $Ba$;
17:     **End if**
18: **End while**

*2) Complexity Analysis*

The computational complexity of the SETG-TG algorithm is dominated by solving obstacle avoidance problems (Lines 3-18), as summarized in Algorithm 1. The SETG-TG algorithm needs to solve $N$ obstacle avoidance problems in the worst case, where $N$ is the total number of obstacles in the environment. Suppose that the computational complexity of one obstacle avoidance (Lines 4-17) is $O(n)$. Hence, in the worst case, the computational complexity of Algorithm 1 is $O(N \times n)$.

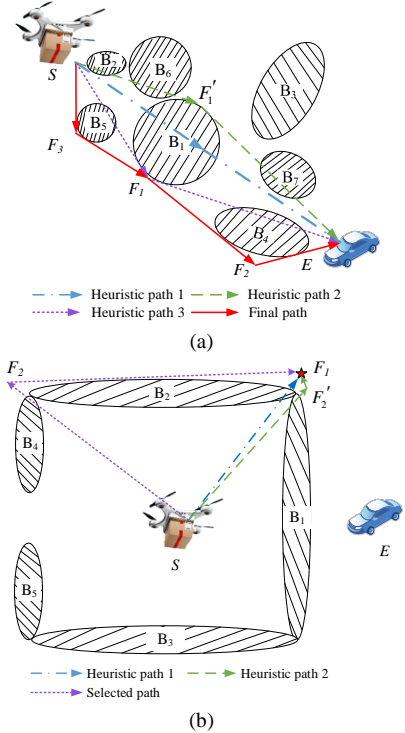

Fig. 3 Path generated by the SETG-TG algorithm. (a) General environment. (b) Maze-like environment.

## B. DETG-TG Algorithm

The DETG-TG algorithm is suitable to either of the following two conditions. The first condition is the dynamic environment with pop-up obstacles. The UAV flies along the offline path generated by the SETG-TG algorithm and meanwhile perceives the environment based on sensors. Once perceived pop-up obstacles collide with the offline path, DETG-TG will re-plan the conflicting sub-path. As shown in Fig. 4. The initial collision-free path, namely $S \to F_1 \to F_2 \to E$, is generated by SETG-TG. When the UAV flies along the initial path, the unexpected obstacle $B_5$ suddenly appears which collides with the offline planned path $F_2E$. At this moment, the DETG-TG algorithm re-plans the conflicting sub-path $F_2E$ and obtains a new path $F_2 \rightarrow F_3 \rightarrow E$ which can successfully avoid the pop-up obstacle $B_5$.

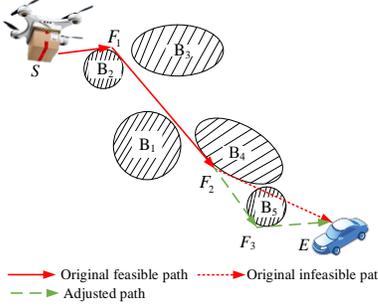

Fig. 4 Path generated by the DETG-TG algorithm in the dynamic environment with pop-up obstacles

The second condition is the completely unknown environment, as shown in Fig. 5. Only a start-point and an end-point are known in advance. The environment information maintained by the planner is initially empty. Based on local information gathered by sensors in real time, the environment information is continuously added and updated at each waypoint. The DETG-TG algorithm generates a collision-free path base on perceived environment step by step like SETG-TG. Compared with the SETG-TG algorithm, the DETG-TG algorithm limits the flight distance between two adjacent waypoints like VFH, which can increase the number of waypoints to perceive the environment frequently based on sensors. Thus, when obtaining a waypoint from which the straight path to an origin is collision-free, we should judge whether the distance between the waypoint and the origin is more than the limited flight distance. As shown in Fig. 5(a), the UAV can perceive the obstacles $B_1$, $B_2$, $B_5$ at the start-point $S$. The waypoint $F_1$ can be generated by SETG-TG. Since the path from the start-point $S$ to the waypoint $F_1$ is collision-free and its length is less than the limited flight distance, the waypoint $F_1$ is feasible. Nevertheless, as shown in Fig. 5(c), the path $F_2F_4$ is collision-free while the length of the path $F_2F_4$ is more than the limited flight distance $l$. Thus, the waypoint $F_4$ is not feasible currently. The waypoint $F_3$ that is $l$ distant from the origin $F_2$ on the path $F_2F_4$ is determined. The pseudocode of the DETG-TG algorithm in completely unknown environments is described in Algorithm 2.

The main steps of the DETG-TG algorithm in completely unknown environments are as follows.

*Step 1 (Initialization):* Add the start-point $S$ and the end-point $E$ to the waypoint set $Pa$ and the set of candidate waypoints $Ca$, respectively (Line 1).

*Step 2 (Update O and D):* Regard the final waypoints of the set $Pa$ and $Ca$ as $O$ and $D$, respectively (Line 3). The UAV adopts various sensors to perceive the flight environment in real time at the point $O$ (Line 4).

*Step 3 (Judging whether the path collides with visible obstacles):* Connect $O$ to $D$ and judge whether the path $OD$ is collision-free (Line 5). If true, go to Step 4; otherwise, mark and record the first-collided visible obstacle and go to Step 5.

*Step 4 (Flying along the path OD):* Judge whether the length of the path $OD$ is more than the limited flying distance $l$. If true, determine the waypoint $T$, which is $l$ away from $O$ in the direction of the path $OD$, and add the waypoint $T$ to the set $Pa$ (Lines 7-8); otherwise, add the point $D$ to the set $Pa$ and delete the point $D$ from the set $Ca$ (Line 10). Go to Step 6.

*Step 5 (Obstacle Avoidance):* Generate two tangents of the first-collided visible obstacle from $O$ and $D$ respectively, thus obtaining two sub-paths (or two intersection points). Then, select the waypoint $T$ according to four rules included in the SETG-TG algorithm and add the waypoint $T$ to the set $Ca$ (Line 13).

*Step 6 (Stop Criteria):* Go to Step 2 and repeat the above steps until the path visits the end-point without obstacle collision (Lines 2-15).

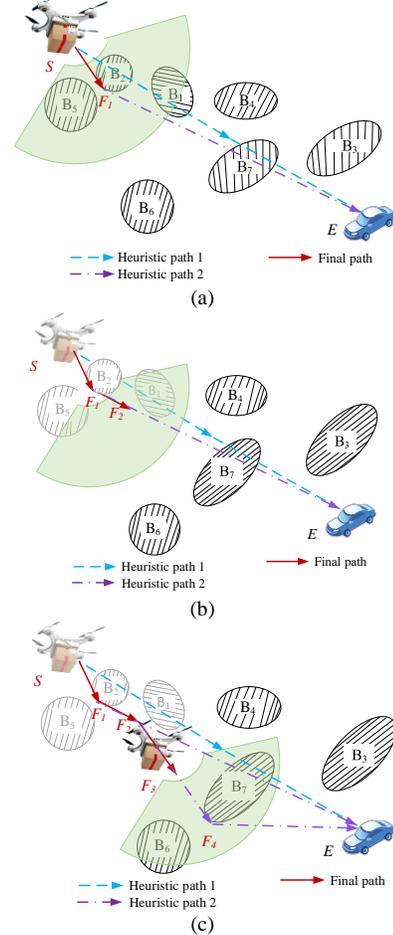

Fig. 5 Path generated by the DETG-TG algorithm in the completely unknown environment. (a) Environment perception at the start-point. (b), (c) Environment perception during the flight.

---

**Algorithm2** DETG-TG

**Input:** Start-point $S$, End-point $E$, the limited flight distance $l$, Search range $R$;

**Output**: $Pa$;

1: Initialize: $Pa \leftarrow \{S\}, Ca \leftarrow \{E\}$;

2: **While** $Ca$ is not empty

3:　　$O \leftarrow Pa(end), D \leftarrow Ca(end)$;

4:　　Update the environment information at the point $O$;

5:　　Connect $O$ and $D$, and judge whether the path $OD$ collides with visible obstacles;

6:　**If** the path $OD$ is collision-free

7:     **If** the length of *OD* is more than *l*
8:         Determine the waypoint *T*, which is *l* away from *O* in the direction of the path *OD*, and add the waypoint *T* to the set *Pa*;
9:     **Else**
10:        Add *D* to *Pa*, and delete *D* from *Ca*;
11:    **End if**
12:    **Else**
13:       Generate two tangents of the first-collided obstacle from *O* and *D* respectively and obtain two sub-paths. Choose a better sub-path, update the waypoint *T* according to four rules and add *T* to *Ca*;
14:    **End if**
15: **End while**

### C. Path Smoothing

The APPATT can generate a series of waypoints and then a winding and collision-free path can be obtained by connecting these waypoints in order. However, this path cannot meet the practical requirements of continuous flight. Thus, the path is smoothed by the cubic B-spline curve [37], [38]. The proposed waypoints are regarded as the control points of B-spline basis function. Then, a smooth path with continuous curvature can be generated. Suppose that the waypoints obtained by APPATT are $\{P_1, P_2, \cdots, P_n\}$. The basis function of the cubic B-spline curve is as follows.

$$\begin{cases} G_{0,3}(t) = \frac{1}{6}(-t^3 + 3t^2 - 3t + 1) \\ G_{1,3}(t) = \frac{1}{6}(3t^3 - 6t^2 + 4) \\ G_{2,3}(t) = \frac{1}{6}(-3t^3 + 3t^2 + 3t + 1) \\ G_{3,3}(t) = \frac{1}{6}t^3 \end{cases} \quad t \in [0,1] \quad (5)$$

The cubic B-spline curve between $P_i$ and $P_{i+3}$ can be formulated as

$$P_{i,i+3} = \frac{1}{6}[1\ t\ t^2\ t^3]\begin{bmatrix} 1 & 4 & 1 & 0 \\ -3 & 0 & 3 & 0 \\ 3 & -6 & 3 & 0 \\ -1 & 3 & -3 & 1 \end{bmatrix}\begin{bmatrix} P_i \\ P_{i+1} \\ P_{i+2} \\ P_{i+3} \end{bmatrix} \quad (6)$$
$$t \in [0,1]$$

## V. COMPUTATIONAL EXPERIMENTS

In this section, we evaluate the performance of APPATT in both static and dynamic environments. The proposed algorithm and the four compared algorithms are coded in MATLAB, and run on a PC computer with Core i5-8400 2.80GHz CPU, 8G memory, and Windows 10 operating system.

### A. General Environments

To obtain reliable results regarding the performance of APPATT to plan paths, the environments in which the experiments will be performed should be as heterogeneous as possible. Thus, five distinct environments [39] are designed as shown in Fig. 6, each of which has unique spatial characteristics. Five environments are marked as E1-E5. As shown in Table II, twenty instances are generated. The simulation scenarios of the instances C1-C10 are applied in the range of 100km×100km, while the simulation scenarios of C11-C20 are applied in the range of 200km×200km.

Five algorithms, namely, A* [17], PRM [24], RRT [25], VFH [9] and SETG-TG are compared in this section. The computational results are reported in Fig. 7 and Table II, where the columns provide the instance, the environment, numbers of obstacles (Num_B), the path length, running time (CPU) in seconds compared with other four algorithms. The instances C7-C10 or C17-C20 are under completely the same environment respectively, while their start-point and end-point are different.

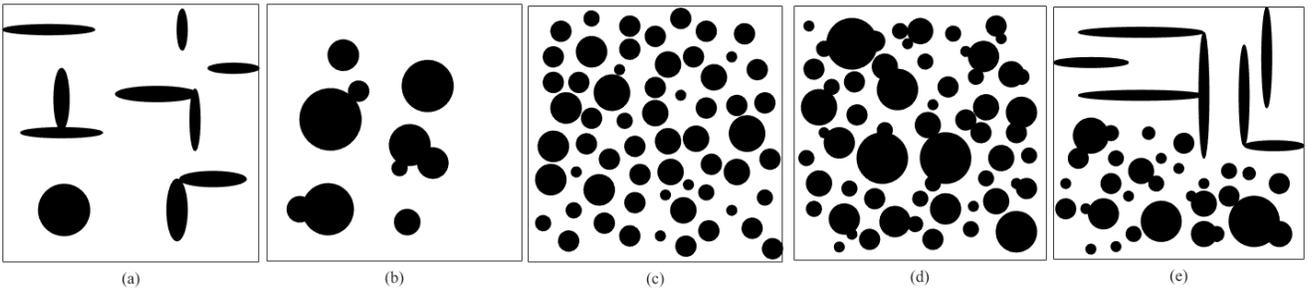

(a)         (b)         (c)         (d)         (e)

Fig. 6 Five distinct environments. (a) E1: Sparse obstacles with corridors. (b) E2: Sparse circular obstacles. (c) E3: Dense non-overlapping circular obstacles. (d) E4: Dense overlapping circular obstacles. (e) E5: Dense overlapping circular obstacles with corridors.

TABLE II

EXPERIMENTAL RESULTS IN THE GENERAL ENVIRONMENTS

| Instance | Env | Num_B | Path length(km) | | | | | CPU(sec) | | | | |
|---|---|---|---|---|---|---|---|---|---|---|---|---|
| | | | A* | PRM | RRT | VFH | SETG-TG | A* | PRM | RRT | VFH | SETG-TG |
| C1 | E1 | 10 | 165 | 139.92 | 184.19 | 138.67 | 126.89 | 0.093 | 6.173 | 0.069 | 0.600 | 0.016 |
| C2 | E2 | 10 | 165 | 136.82 | 169.85 | 138.40 | 131.87 | 0.063 | 4.746 | 0.044 | 1.017 | 0.002 |
| C3 | E3 | 60 | 165 | 184.42 | 137.77 | 153.23 | 125.34 | 0.054 | 8.823 | 0.121 | 1.017 | 0.011 |
| C4 | E4 | 60 | 173 | 180.35 | 159.30 | 162.72 | 126.47 | 0.051 | 9.188 | 0.113 | 1.269 | 0.019 |
| C5 | E3 | 80 | 165 | 189.73 | 157.01 | 128.79 | 131.55 | 0.063 | 8.367 | 0.075 | 1.081 | 0.039 |
| C6 | E4 | 80 | 167 | 185.69 | 165.58 | 155.65 | 125.24 | 0.056 | 9.008 | 0.287 | 1.052 | 0.033 |
| C7 | E5 | 38 | 217 | 197.38 | 283.66 | 203.81 | 178.34 | 0.084 | 5.770 | 0.487 | 1.340 | 0.015 |
| C8 | E5 | 38 | 217 | 198.69 | 211.37 | 199.61 | 175.82 | 0.063 | 5.434 | 0.240 | 1.651 | 0.014 |

| | | | | | | | | | | | |
|---|---|---|---|---|---|---|---|---|---|---|---|
| C9  | E5 | 38  | 154 | 251.03 | 119.93 | 149.50 | 133.45 | 0.047 | 4.872  | 0.227 | 1.293 | 0.014 |
| C10 | E5 | 38  | 194 | 186.38 | 211.85 | 237.55 | 177.39 | 0.061 | 5.155  | 0.447 | 1.836 | 0.010 |
| C11 | E1 | 20  | 365 | 296.98 | 366.49 | 362.31 | 286.83 | 0.639 | 4.881  | 0.074 | 1.655 | 0.002 |
| C12 | E2 | 20  | 365 | 309.70 | 403.21 | 340.75 | 310.79 | 0.495 | 4.394  | 0.102 | 1.880 | 0.004 |
| C13 | E3 | 120 | 367 | 370.28 | 302.13 | 350.64 | 274.88 | 0.343 | 8.963  | 0.161 | 1.858 | 0.022 |
| C14 | E4 | 120 | 365 | 331.37 | 412.21 | 350.11 | 301.10 | 0.390 | 9.013  | 0.457 | 1.782 | 0.034 |
| C15 | E3 | 150 | 365 | 365.53 | 341.43 | 386.55 | 275.02 | 0.481 | 10.477 | 0.838 | 2.234 | 0.050 |
| C16 | E4 | 150 | 365 | 350.89 | 332.69 | 426.19 | 303.50 | 0.409 | 8.623  | 0.406 | 2.503 | 0.054 |
| C17 | E5 | 40  | 365 | 294.12 | 420.76 | 618.52 | 278.84 | 0.464 | 5.886  | 0.857 | 2.478 | 0.018 |
| C18 | E5 | 40  | 365 | 292.65 | 368.02 | 305.08 | 277.39 | 0.459 | 5.932  | 0.694 | 1.507 | 0.008 |
| C19 | E5 | 40  | 344 | 326.74 | 466.63 | 390.58 | 313.53 | 0.460 | 5.908  | 0.463 | 1.949 | 0.018 |
| C20 | E5 | 40  | 190 | 166.80 | 359.33 | 416.23 | 155.90 | 0.173 | 6.614  | 0.827 | 2.054 | 0.012 |

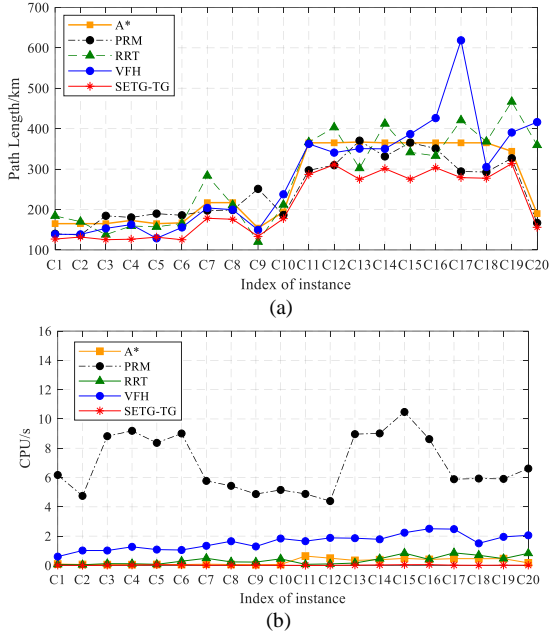

Fig. 7 Comparisons of the path length and the time consumption. (a) Path length for each instance. (b) Running time for each instance

From Fig.7 (a), it can be observed that the SETG-TG algorithm always generates the shortest path for each instance among the algorithms, demonstrating that SETG-TG has a strong capability of path optimization in the general environments. This is because estimated path lengths and obstacle avoidance conditions of destination-tangents are taken into account. Due to the waypoint generation based on randomized sampling, classical PRM and RRT cannot guarantee the path optimality and may easily cause the UAV to take a detour. By contrast, the A* algorithm is a deterministic algorithm based on grids. The size of each grid in the experiments is 1km × 1km. There is no doubt that the A* algorithm is capable of escaping from simple traps like the environment E5 as shown in Fig. 6(e). However, the quality of the generated path for each instance is not satisfactory. The VFH algorithm has a satisfactory performance in C1-C9, whereas it generally obtains the worst path in C15-C20. This is because VFH is a local path planning algorithm with the limited search range, which can only avoid visible obstacles. Moreover, classical VFH prefers wide channels so that it is easy to fall into the local optima when facing corridors in the environment E5.

Among the five algorithms, SETG-TG always generates the collision-free path requiring the shortest time within 0.05 seconds for each instance, followed by A* and RRT. The reason about high time efficiency of SETG-TG is that it mainly focuses on the obstacles on the path from the UAV's location to the target and navigates the UAV to the target by the elliptic tangent graph method, which can avoid unnecessary detours. Since VFH is a local path planning algorithm which looks for gaps with the limited search range, it consumes nearly 2 seconds in the static environment for each instance. The PRM algorithm is computationally intensive especially in densely clustered environments because it has many two-point boundary value problems. Table II also shows that the more complex the environment is, the more obvious the superiority of SETG-TG in time consumption is.

SETG-TG and VFH have the common point that they both select promising directions continuously to navigate the UAV to the target without obstacle collision. However, they differ in the obstacle avoidance strategy and the reference direction selection strategy. The VFH algorithm selects a promising direction every time by looking for gaps in locally constructed polar histograms. Nevertheless, SETG-TG selects a promising direction via the elliptic tangent graph method. Strictly speaking, the paths generated by SETG-TG could be closer to obstacles compared with VFH. Owing to the construction of a polar histogram continuously, VFH consumes more time than SETG-TG, especially in densely obstructive environments. In addition, VFH always selects the direction from the UAV's momentary location to the target as the reference direction, whereas SETG-TG sometimes regards the direction from the UAV's location to the waypoint near the target as the reference direction. In other words, the destination in VFH is always the end-point whereas the destination in SETG-TG is a variable point which can be updated in real time. For example, in Fig. 3(b), the initial destination in SETG-TG is the end-point $E$. Next time, SETG-TG regards the waypoint $F_1$ as the destination. On the contrary, VFH always takes the end-point $E$ as its destination.

To sum up, the SETG-TG algorithm shows a superior path planning performance compared to other four algorithms in terms of path length and time consumption in the static environments.

*B. Simple Maze-like Environments*

Maze-like flying environments exist in some complex application scenarios. Hence, it is necessary to investigate the path planning performance of the SETG-TG algorithm in simple maze-like environments. Six instances called C21-C26 are generated as shown in Fig. 8. The results are reported in Table III. From Fig. 8, SETG-TG succeeds in escaping from traps and generating a collision-free path for each instance. The reference direction selection strategy enables the UAV to escape from traps. However, the paths may be tortuous owing to the above-mentioned strategy. For example, in Fig. 8(e), the direction from the start-point $S$ to the end-point $E$ is regarded as the reference direction that collides with the obstacle $B_1$ represented by a yellow ellipse. Two sub-paths are generated by drawing two tangents of the obstacle $B_1$ from the point $S$ and $E$ respectively. One of the two sub-paths that passes the point $F$ is selected according to heuristic rules in the first obstacle avoidance. Then the direction of the path $SF$ is regarded as the reference direction in the next obstacle avoidance. The waypoint $F$ contributes to escaping from traps while it leads to detour as well. More significantly, Table III shows that SETG-TG generally consumes less than 0.1s (except for the instance C26) to find a collision-free path in the considered simple mazes.

TABLE III
EXPERIMENTAL RESULTS IN THE MAZE-LIKE ENVIRONMENTS

| Instance | Path length(km) | CPU(sec) |
|---|---|---|
| C21 | 154.50 | 0.026 |
| C22 | 168.81 | 0.041 |
| C23 | 106.26 | 0.014 |
| C24 | 271.02 | 0.022 |
| C25 | 235.18 | 0.061 |
| C26 | 275.10 | 0.108 |

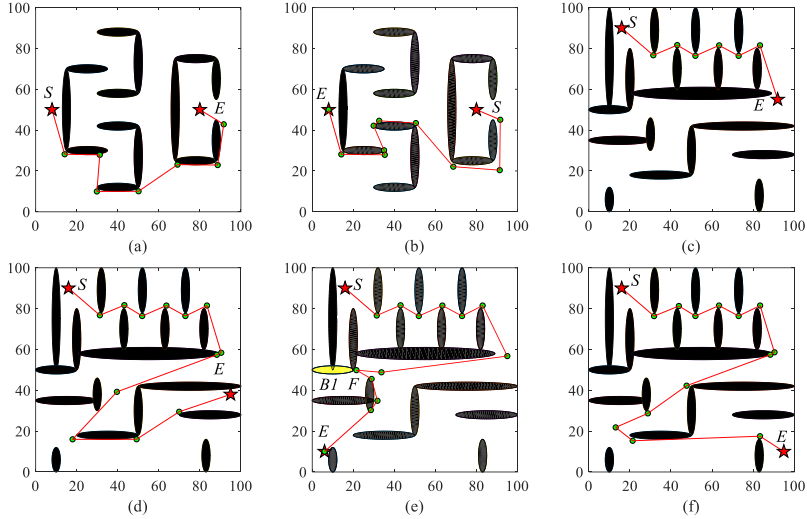

Fig. 8 Paths generated by the SETG-TG algorithm for the simple maze-like environments. (a) - (f): C21-C26.

## C. Dynamic Environments

To test the effectiveness of the proposed algorithm in dynamic environments, we conduct simulation experiments in the partly or completely unknown environments. The maximum flight range of the UAV is 300 kilometers. The limited flight distance of the UAV is 3 kilometers. The search range of the UAV is 10 kilometers. The minimum turning radius is 0.2 kilometers.

**Simulation experiment I:** path planning in the dynamic environment with pop-up obstacles

We prepare three instances (i.e., C27-C29) to compare DETG-TG with VFH in the dynamic environments with pop-up obstacles as shown in Fig. 9. Compared with VFH, DETG-TG can generate a shorter and smoother path. In addition, DETG-TG consumes less time to re-plan path when facing pop-up obstacles. Particularly, the running time of path re-planning is cut down by 90% and the path length is shortened by 4% in comparison to VFH. Nevertheless, the paths generated by DETG-TG are closer to obstacles since a promising direction is selected between two tangents generally considering the safe distance. As shown in Fig. 9(c) and (d), there is a narrow channel between two pop-up obstacles. The difference on the obstacle avoidance strategy could be verified from the fact that DETG-TG prefers to select the narrow channel while VFH would select the wide channel.

**Simulation experiment II:** path planning in completely unknown environments

As shown in Fig. 10, three instances (C30-C32) in completely unknown environments are generated. The DETG-TG algorithm is tested by using above instances and compared with VFH and RRT based on rolling windows (RRTRW) [40]. The results are reported in Fig. 10 and Table IV.

The table shows that the DETG-TG algorithm always produces the shortest path for each instance among algorithms. In general, RRTRW obtains the longest path of each instance because RRTRW generates the roadmap in a random manner without the guarantee of path optimality. In contrast, VFH is superior to RRTRW but not as competitive as the DETG-TG with regard to the path length. Regarding time efficiency, RRTRW is the most efficient, and DETG-TG consumes less time than VFH.

TABLE IV
EXPERIMENTAL RESULTS IN COMPLETELY UNKNOWN ENVIRONMENTS

| Instance | Path length(km) | | | CPU(sec) | | |
|---|---|---|---|---|---|---|
| | RRTRW | VFH | DETG-TG | RRTRW | VFH | DETG-TG |
| C30 | 177.47 | 154.36 | 132.98 | 2.79 | 6.24 | 6.22 |
| C31 | 222.63 | 147.11 | 135.04 | 4.40 | 10.72 | 7.13 |
| C32 | 278.82 | 147.06 | 114.02 | 3.48 | 7.59 | 1.68 |

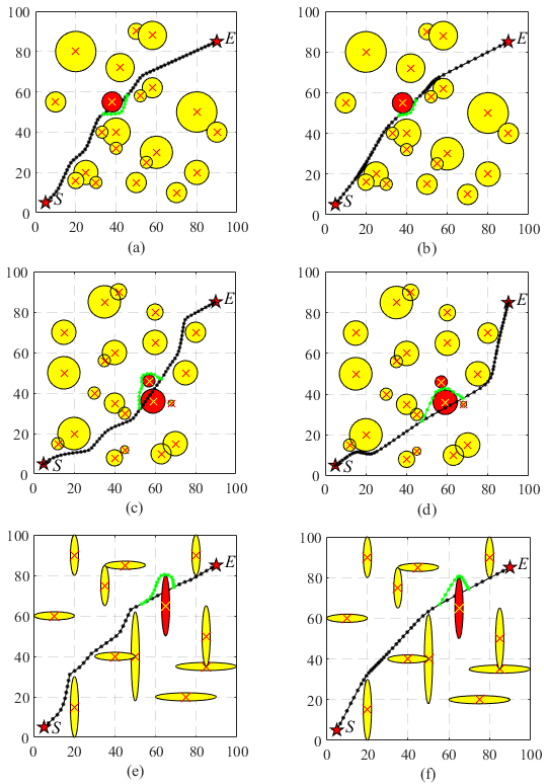

(e) are paths generated by VFH. (b), (d) and (f) are paths generated by the DETG-TG. The yellow ellipses denote known obstacles. The red ellipses denote pop-up obstacles. The red stars denote the start-point and end-point. The black line and green line denote the initial path and re-planned path, respectively.

In summary, the results demonstrate that the proposed DETG-TG algorithm could respond rapidly to the pop-up obstacles and re-plan the conflicting sub-path effectively. Furthermore, the DETG-TG algorithm could plan a collision-free path in real time when facing the completely unknown environments.

## VI. Conclusion

In this paper, to generate collision-free and high-quality paths in static or dynamic environments, we propose a novel graph-based algorithm for UAVs, namely APPATT. Guided by a target, when confronting an obstacle, two sub-paths are generated by the elliptic tangent graph method and meanwhile one of the two sub-paths is selected based on four heuristic rules. The operation is iteratively performed until the collision-free path extends to the target. The APPATT has two versions, one is the SETG-TG for static environments and the other one is the DETG-TG for dynamic environments. The effectiveness of APPATT is validated by extensive computational experiments, and the following conclusions are obtained.

Fig. 9 Paths in the dynamic environments with pop-up obstacles. (a), (c) and

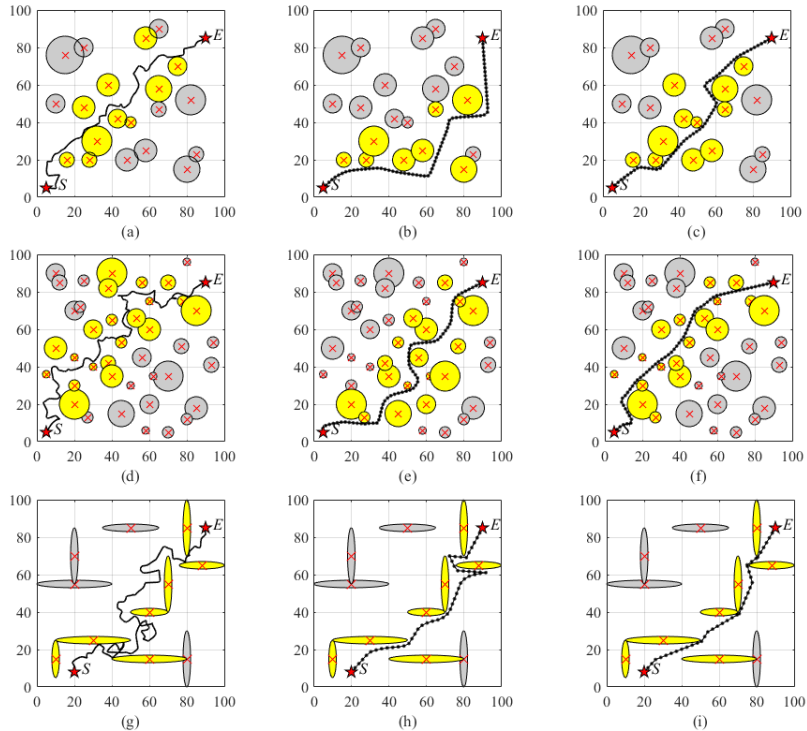

Fig. 10 Paths in the completely unknown environments. (a), (d), (g) RRTRW in C30-C32. (b), (e), (h) VFH in C30-C32. (c), (f), (i) DETG-TG in C30-C32. The yellow ellipses denote visible obstacles. The grey ellipses denote invisible obstacles. The red stars denote the start-point and end-point. The black line denotes the final path.

First, the SETG-TG algorithm can generate satisfactory, feasible and collision-free paths with high time efficiency in the static environments. In particular, the SETG-TG algorithm just consumes 0.05 seconds under different instances with dense obstacles, which is far less than the time consumption of other compared algorithms. Second, the SETG-TG algorithm has the capability of escaping from traps in simple maze-like environments. Third, the DETG-TG algorithm is a local search algorithm, which can rapidly respond to pop-up obstacles and plan a collision-free path in real time without the environment information in advance.

The limitations of APPATT mainly lie in two aspects. First, it is a deterministic algorithm, which generates collision-free paths in a stepwise manner based on some sophisticatedly designed rules. Although the experiments and comparison studies demonstrate the outstanding performance of APPATT, currently no theory guarantees its optimality. Second, APPATT can achieve the goal of producing high-quality collision-free paths for UAVs and experiments show that it works in simple maze environments. Nevertheless, there is no theory guarantee that APPATT can help UAVs to fly to a target successfully in complex maze environments. Thus, future studies will aim to design more sophisticated rules to improve the quality of path planning, and the possibility of finding satisfactory solutions in complex maze environments.